\def\year{2021}\relax
\documentclass[letterpaper]{article} 
\usepackage[linesnumbered,ruled,vlined]{algorithm2e}
\usepackage{algpseudocode}
\usepackage{amsmath}
\usepackage{color}
\usepackage{amsfonts,amssymb}
\usepackage{aaai21}  
\usepackage{times}  
\usepackage{helvet} 
\usepackage{courier}  
\usepackage[hyphens]{url}  
\usepackage{graphicx} 
\usepackage{natbib}
\usepackage{caption} 
\title{Consensus Graph Representation Learning  for Better \\ Grounded Image Captioning}
\author {
Wenqiao  Zhang\textsuperscript{\rm 1}\footnote{These authors contributed equally to this work.},
        Haochen Shi \textsuperscript{\rm 1}\footnotemark[\value{footnote}],
        Siliang Tang \textsuperscript{\rm 1}\footnote{Corresponding author.},
        Jun Xiao \textsuperscript{\rm 1},
        Qiang Yu \textsuperscript{\rm 2},
       Yueting Zhuang \textsuperscript{\rm 1}\\
}
\affiliations {
    \textsuperscript{\rm 1} Zhejiang University,
    \textsuperscript{\rm 2} Citycloud Technology\\
    wenqiaozhang,hcshi,siliang,junx,yzhuang@zju.edu.cn,  yq@citycloud.com.cn
}

\graphicspath{{pic/}
}

\begin{document}

\maketitle

\begin{abstract}
The contemporary visual captioning models frequently hallucinate objects that are not actually in a scene, due to the visual misclassification or over-reliance on priors that resulting in the semantic inconsistency between the visual information and the target lexical words. The most common way is to encourage the captioning model to dynamically link generated object words or phrases to appropriate regions of the image, i.e., the grounded image captioning (GIC). However, GIC utilizes an auxiliary task (grounding objects) that has not solved the key issue of object hallucination, i.e., the semantic inconsistency. In this paper, we take a novel perspective on the issue above - exploiting the semantic coherency between the visual and language modalities. Specifically, we propose the Consensus Graph Representation Learning framework (CGRL) for GIC that incorporates a consensus representation into the grounded captioning pipeline. The consensus is learned by aligning the visual graph (\emph{e.g., scene graph}) to the language graph that   consider both the nodes and edges in a graph. With the aligned consensus, the captioning model can capture both the correct linguistic characteristics and visual relevance, and then grounding appropriate image regions further. We validate the effectiveness of our model, with a significant decline in object hallucination (-9\% CHAIR$_i$) on the Flickr30k
Entities dataset. Besides, our CGRL also evaluated by several automatic metrics and human evaluation, the results indicate that the proposed approach can simultaneously improve the performance of image captioning (+2.9 Cider) and  grounding (+2.3 F1$_{LOC}$).
\end{abstract}
\section{Introduction}

The ability of understanding and reasoning different
modalities (e.g., image and language) is
a longstanding and challenging goal of artificial intelligence~\cite{rennie2017self,wang2019multilayer,he2019vd,zhang2019frame,zhang2020photo,antol2015vqa}. Recently, image captioning models have achieved impressive or even super-human performance on many benchmark datasets~\cite{shuster2019engaging,deshpande2019fast,zhang2021tell}. However, further quantitative analyses show that they are likely to generate hallucinated captions~\cite{zhou2019grounded,ma2019learning}, such as hallucinated objects words.  Previous studies~\cite{rohrbach2018object} believed that this caption hallucination problem is caused by biased or inappropriate visual-textual correlations learned from the datasets, i.e., the semantic inconsistency between the visual and language domain. Therefore, Grounded Image Captioning  (GIC) is proposed to tackle this problem by introducing a new auxiliary task that requires the captioning model to ground object words back to the corresponding image regions during the caption generation. The auxiliary grounding task provides extra labels between visual and textual modals that can be used to remove biases and to reform the correct correlations between the two modalities.

 However, GIC may not be the true savior of this hallucination problem. First, grounding object words is still far from solving the problem since the model can still hallucinate attributes of the objects and also relationships among the objects. Of course, we can introduce more grounding tasks to alleviate these new problems, but it comes at a tremendous cost and may induce new biases that are even harder to detect. Second, the correct correlations can hardly be fully constructed by grounding the annotation, since the image and the annotated caption are not always consistent~\cite{yang2020deconfounded}. As is commonly known, such inconsistency happens frequently in real life, however, we humans have the reasoning ability to summarize or to infer the consensus knowledge between currently imperfect information and existing experience from the inconsistent environment. This ability enables us to perform better than machines in high-level reasoning and would be the most precious capacity for modern AI. Therefore, it is more important to enhance the reasoning capacities of models rather than just create more annotations.

Based on this insight, we propose a novel learning framework to mimic the human inference procedure, i.e., consensus graph representation learning framework (CGRL). The CGRL can leverage structural knowledge (\emph{e.g. scene graph, $\mathcal{SG}$ }) from both vision and text, and further generate grounded captions based on consensus graph reasoning to alleviate the hallucination issue. As shown in Figure~1,  the consensus representation is inferred by aligning visual graph to language graph. Utilizing such consensus, the model can capture the accurate and fine-grained information among objects to predict the non-visual function words reasonably (\emph{e.g.,  relationship verbs: ``ride'', ``play'',  attribute adjective: ``red'', `` striped''}), while these words are inherently challenging to predict in GIC manner. Besides,  the appropriate correlations between image and the annotated caption can be mined in the consensus to against the semantic inconsistency across the visual-language domain. Exactly as illustrated in Figure~1, although visual and language graphs are quite varied, the CGRL employs the aligned consensus that can maintain semantic consistency, and then generates the accurate description which captures both the correct linguistic characteristics and visual relevance. In addition, the object words ``woman'', ``shirt'', ``laptop'' and ``table'' are also grounded appropriately on the spatial regions.

\begin{figure}[t]
\includegraphics[width=0.5\textwidth]{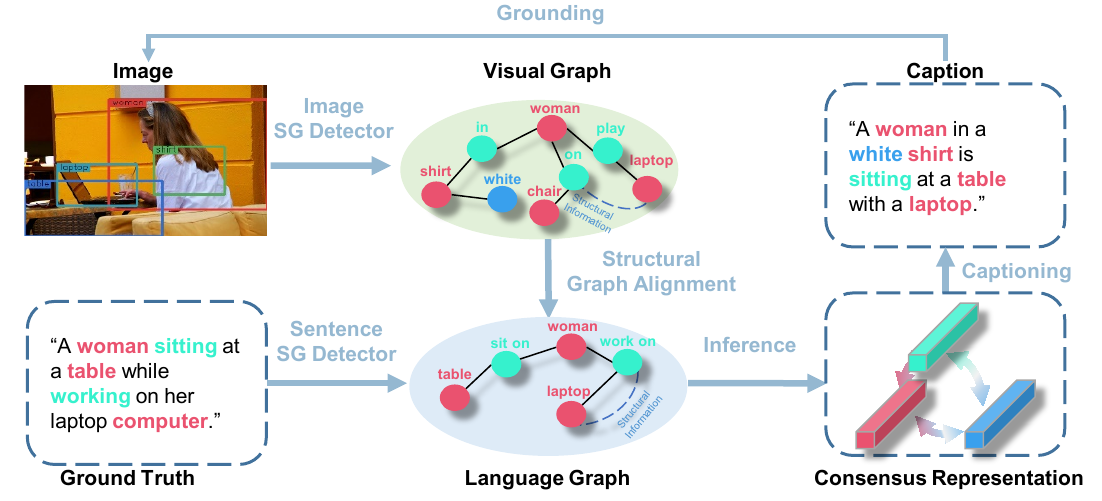}
\centering\caption{ An example of how consensus facilitates the grounded image captioning.  Pink, green and blue
 represent the objects, relationships and attributes, respectively.}
\vspace{-0.5cm}
\end{figure}

Specifically, in our setting, the training pipeline of CGRL consists of three parts: 1) \emph{Consensus Reasoning Module}, we first build the image scene graph $\mathcal{SG^{\mathcal{V}}}$ and the language scene graph $\mathcal{SG^{\mathcal{L}}}$ from the image and its ground truth (GT) at the training stage.  Then we infer the consensus representation by aligning the $\mathcal{SG^{\mathcal{V}}}$  to $\mathcal{SG^{\mathcal{L}}}$. This is a challenging task: as shown in Figure~1, the category and number of visual concepts in $\mathcal{SG}$ are varied in visual and language domain; besides, for a graph, the structural information (\emph{e.g., the diversity among objects, attributes and relationships}) also needs to be aligned. To infer the consensus, we propose a Generative Adversarial Structure Network (GASN) that aligns the $\mathcal{SG^{\mathcal{V}}}$  to $\mathcal{SG^{\mathcal{L}}}$. We first encode the $\mathcal{SG}$ at three unified levels (object, relationship, attribute) by GCN, then we simultaneously align the nodes and edges in the encoded graph by GASN to exploit the semantic consistency across domains. The representation of the aligned result can be regarded as the consensus for better GIC, 2) \emph{Sentence Decoder},  we first exploit the latent spatial relations between image proposals and link them as the augmented region information for the sentence generator. Then the sentence decoder learns to decide the utilization of the augmented regions and consensus representation to describe an image more reasonably and accurately.   3)  \emph{Grounding Module}, we build a grounding and localizing mechanism. It not only encourages the model to dynamically ground the regions based on the current semantic context to predict words, but also localizes regions by the generated object words. Such a setting can boost the accuracy of the object word generation.

To summarize, the major contributions of our paper are as follows: 1)We present a novel perspective that alleviates the issue of object hallucination (CHAIR$_i$: -9\%) by utilizing the consensus to maintain the semantic consistency across vision-language modality; 2)We propose a novel Consensus Graph Representation Learning framework (CGRL) that organically incorporates the consensus representation into the GIC pipeline for better grounded caption generation; 3)We propose an adversarial graph alignment method to reason the consensus representation, which mainly addresses the problem of data representation and structural information for graph alignment;
4)We demonstrate the superiority of CGRL by automatic metrics and human judgments in boosting the captioning quality (Cider: +2.9)and grounding accuracy (F1$_{LOC}$: +2.3) over the state-of-the-art baselines.

\section{Related Work}
\noindent$\textbf{Image Captioning.}$ Image captioning has been actively studied in recent vision and language research, the prevailing
image captioning techniques often incorporate the encoder-decoder pipeline inspired by the first successful model~\cite{vinyals2015show}. Benefiting from the rapid development of deep learning, the image captioning models have achieved striking advances by attention mechanism~\cite{wang2019multilayer}, reinforcement learning~\cite{rennie2017self,zhang2017actor} and generative adversarial networks~\cite{chen2017show,xu2019exact}. Although these methods have reached impressive performance on automatic metrics, they usually neglect how well the generated caption words are grounded in the image, making models less explainable and trustworthy.

\noindent$\textbf{Visual Grounding.}$ Visual grounding models encourage captioning generator link phrases to specific spatial regions of the images, which present a potential way to improve the explainability of models. The most common way adopted by grounding models~\cite{rohrbach2016grounding,xiao2017weakly,zhou2019grounded,zhang2020relational} is to predict the next word through an attention mechanism, which is deployed over the NPs (Noun Phrases) with supervised bounding boxes as input. However, in GIC, visual grounding as an auxiliary task that has not truly solve the problem of semantic inconsistency can still hallucinate objects in the generated caption.

\noindent$\textbf{Scene Graph.}$ Recently, $\mathcal{SG}$ construction has become popular research topics with significant advancements  ~\cite{zellers2018neural,yang2018graph,gu2019scene,zhang2017visual} based on the Visual Genome~\cite{krishna2017visual} dataset.  The  $\mathcal{SG}$ contains the structured semantic information, it can represent scenes as directed graphs. Using this priors knowledge is natural to improve the performance of vision-language task, e,g., image captioning~\cite{yang2019auto,gu2019unpaired}, VQA~\cite{shi2019explainable,teney2017graph}. However, directly fed the $\mathcal{SG}$ to the captioning model may lead to the non-correspondence problem between vision and language. Thus, how to infer the consensus knowledge from $\mathcal{SG}$ is the key to promote the vision-language field further.

\begin{figure*}[t]
\includegraphics[width=0.9\textwidth]{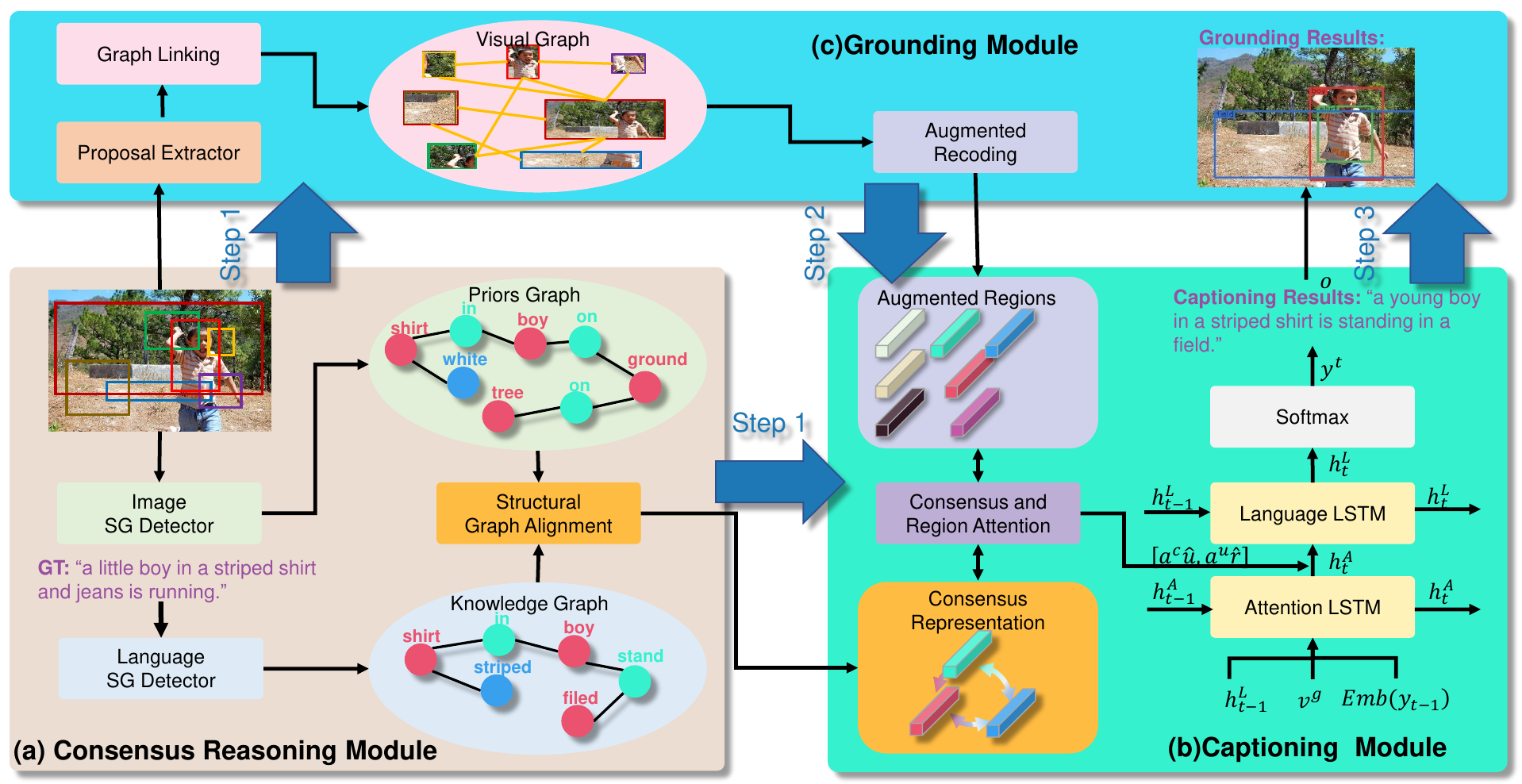}
\centering\caption{Overview of our method. It consists of three modules: (a)\emph{Consensus  Reasoning Module} first transforms an image and its GT into the $\mathcal{SG}$. Then we develop a Generative Adversarial Structure Network (GASN)  that aligns the graphs from the visual domain to the language domain and learn the consensus representation.  (Section 3.2);
 \emph{Captioning Module} is encouraged to predict the textual word based on the consensus and augmented regions. (Section 3.3); \emph{Grounding Module} compute the accuracy of localization of object words on the GT sentence, and the captioning model dynamically attends region proposals in the word prediction stage. (Section 3.4). The blue arrow represents the order of interaction between modules.}
\vspace{-0.4cm}
\end{figure*}
\section{Method}
\subsection{Task Description}
Before presenting our method, we first introduce some basic notions and terminologies. Given an image $\mathcal{I}$, the goal of grounded image captioning (GIC) is to generate the natural language sentence $\mathcal{S}$, and localizing the object words with certain regions $\mathcal{R}^o$=$\{  r_1^o, r_2^o, $ $\cdots, r_k^o$\}  in the corresponding image, where \emph{K} is the number of grounding words appearing in sentence. Let $\mathcal{M}$ be the GIC model with initialized parameter $\Theta$, trained to generate a sentence $\mathcal{\hat{S}}$ and localize the object regions $\mathcal{\hat{R} }^o$ for an image. \emph{i.e.}, ($\mathcal{\hat{S}}$, $\mathcal{\hat{R}}^o$ )= $\mathcal{M}$($\mathcal{I}$).  We define the loss for a training pair as $\mathcal{L}$(($\mathcal{S}$, $\mathcal{R}^o$ ), $\mathcal{M}$($\mathcal{I}$; $\Theta$)). Fig.2 illustrates the overview of our  Consensus Graph Representation Learning(CGRL) system, we illustrate our method for the GIC task as an example for clarity.

\subsection{Consensus Reasoning Module}
In our setting, we regard the pre-extracted $\mathcal{SG}$ from an image and its corresponding GT sentence as the visual graph $\mathcal{SG}^\mathcal{V}$ and the language graph $\mathcal{SG}^\mathcal{L}$, respectively. We aim to translate the $\mathcal{SG}^\mathcal{V}$ to $\mathcal{SG}^\mathcal{L}$, the result of translation  can be regarded as the consensus knowledge. Generally, the $\mathcal{SG} = (\mathcal{N}, \mathcal{E})$ defined in our task contains a set of nodes $\mathcal{N}$ and  edges $\mathcal{E}$. The node set $\mathcal{N}$ contains three types of nodes: object node $o$, attribute node $a$, and relationship node $r$.

\subsubsection{Scene Graph Encoder}
We first follow ~\cite{yang2019auto} and ~\cite{schuster2015generating} to extract the $\mathcal{SG}^\mathcal{V}$  and $\mathcal{SG}^\mathcal{L}$, respectively.
To represent the nodes of $\mathcal{SG}$ at a unified level,  we first denote the nodes $\{o, a, r\}$ by the label embeddings $\{\textbf{e}_o, \textbf{e}_a, \textbf{e}_r\}$ $\in \mathbb{R}^{e}$, corresponding respectively to objects, attributes and relationships both in  $\mathcal{SG}^\mathcal{V}$  and  $\mathcal{SG}^\mathcal{L}$ . Then we employ node encoder of the Graph Convolutional Network (\emph{GCN})~\cite{marcheggiani2017encoding,yang2019auto} to encode the three nodes in $\mathcal{SG}$ at a unified representation $\mathcal{U}=\{\textbf{u}_o, \textbf{u}_a, \textbf{u}_r\}$ $\in \mathbb{R}^{u}$. Note that all the \emph{GCNs} are defined with the same structure but independent parameters.

\noindent$\textbf{Objects Encoding.}$ For an object $o_i$ in $\mathcal{SG}$, it can act different roles (``\emph{subject}'' or ``\emph{object}'') due to different edge directions, \emph{i.e.}, two triples with different relationships, $\langle {o_i} - r_{i,j} - {o_j} \rangle$ and $\langle o_k - r_{k,i} - o_i \rangle$. Such associated objects by the cascaded encoding scheme can represent the global information of objects for an image or a sentence. Therefore we compute $\textbf{u}_{o}$ by explicit role modeling:
\begin{equation}\label{2}
\begin{aligned}
\textbf{u}_{o} =& \sum_{i =1}^{N_o}(a_i^o \cdot \frac{1}{N_{o_i}} (\sum_{\textbf{e}_{o_i} \in sub} E_{subject}(\textbf{e}_{o_i}, \textbf{e}_{r_{i,j}} , \textbf{e}_{o_j} )
\\ & + \sum_{\textbf{e}_{o_i} \in obj } E_{object}(\textbf{e}_{o_k}, \textbf{e}_{r_{k,i}}, \textbf{e}_{o_i} ) ) )
\end{aligned}
\end{equation}
where $N_{o_i}$ = $ \left|sub(o_i) \right| +\left|obj(o_i) \right|$ is the total number of the relationship triplets for object $o_i$. $N_o$ is the number of objects. $a_i^o$ is the soft attention computed by all the objects. $E_{subject}(.)$ and $E_{object}(.)$ are the graph convolutional operation for objects as a ``\emph{subject}'' or an ``\emph{object}''.

\noindent$\textbf{Attributes Encoding.}$ Given $o_i$ in $\mathcal{SG}$, it usually includes several attributes \{$a_1^i, \cdots, a^i_{N_{a_i}} $\}, where $N_{a_i}$ is the total number of its attributes. Therefore, $\textbf{u}_{a}$ can be computed as:
\begin{equation}\label{3}
\begin{aligned}
\textbf{u}_{a} =& \sum_{i=1}^{N_a}( a_i^a \cdot \frac{1}{N_{a_i}} \sum_{i \in N_{a_i}} E_{attribute}(\textbf{e}_{o_i}, \textbf{e}_{a_{i}})))
\end{aligned}
\end{equation}
where $E_{attribute}(.)$ is the graph convolutional operation for object $o_i$ and its attributes. $N_a$ is the number of attributes. $a_i^a$ is the soft attention computed by all the attributes.

\noindent$\textbf{Relationships Encoding.}$  Between two salient objects $o_i$ and $o_j$, their relation is given by the triplet $\langle o_i - r_{i,j} - o_j\rangle$. Similarly, the relationship encoding $\textbf{u}_{r}$ is produced by:
\begin{equation}\label{4}
\begin{aligned}
\textbf{u}_{r} =& \sum_{i =1}^{N_{r_i}} \sum_{j =1}^{N_{r_j}}( a_i^r \cdot E_{relationship}(\textbf{e}_{o_i}, \textbf{e}_{r_{i,j}}, \textbf{e}_{o_j} ))
\end{aligned}
\end{equation}
where $E_{relationship}(.)$ is the graph convolutional operation for relational object $o_i$ and $o_j$. ${N_{r_i}}$ and ${N_{r_j}}$ is the number of relationships for object $o_i$ and $o_j$. $a_i^r$ is the soft attention computed by all the relationships.

After graph encoding, for each $\mathcal{SG^V}$  and  $\mathcal{SG^L}$ , we have three node embeddings:

\begin{equation}
\begin{aligned}
\mathcal{U} ^{\mathcal{M}} =\{{\textbf{u}_o^{\mathcal{M}},\textbf{u}_a^{\mathcal{M}},\textbf{u}_r^{\mathcal{M}}}\},  \mathcal{M} \in \{\mathcal{V},\mathcal{L}\}
\end{aligned}
\end{equation}

\subsubsection{Adversarial Consensus Representation Generation}
To summarize and infer the consensus representation to assist the model for better GIC, we need to translate the $\mathcal{SG^V}$ to $\mathcal{SG^L}$.  Instead of directly modeling the distribution alignment across domains, we take the discrepancy in the modality of $\mathcal{SG}$ directly into account by aligning the data representation and structural information of $\mathcal{SG^V}$ with the $\mathcal{SG^L}$.

In our setting, we link the  $\mathcal{U}$  to build a complete graph $\mathcal{G}_c = (\mathcal{N}_c, \mathcal{E}_c)$, where nodes are $\mathcal{N}_c$= \{$\textbf{u}_o$,$\textbf{u}_a$,$\textbf{u}_r$\} and edge is the distance between each pair of nodes. The edge values $\mathcal{E}_c$ are calculated based on cosine similarity. For example, $\textbf{e}_{c_{oa}}$=$exp (- \frac{\vec{\textbf{u}_o}\cdot \vec{\textbf{u}_a} }{\left \| \textbf{u}_o \right \|\left \| \textbf{u}_a \right \|})$ is the value of edge between nodes $\textbf{u}_o$ and $\textbf{u}_a$. To reason the consensus, we develop a Generative Adversarial Structure Network (GASN) to align the nodes and edges from  $\mathcal{SG^V}$ to $\mathcal{SG^L}$. As shown in Figure~3, the translator $T_{V\rightarrow L}(\cdot)$ is trained to generate the ($\tilde{\mathcal{N}}_c,\tilde{\mathcal{E}}_c$),  the overall loss $\mathcal{L}_G$ of GASN
consists of two parts:
\begin{equation}
\begin{aligned}
\mathcal{L}_G&=\lambda_N \cdot \mathcal{L}(N)+ \lambda_E \cdot \mathcal{L}(E)
\end{aligned}
\end{equation}
where $\mathcal{L}(N)$ and $\mathcal{L}(E)$ are the cross-entropy losses for the node alignment and edge alignment, respectively.
\begin{figure}[t]
\includegraphics[width=0.47\textwidth]{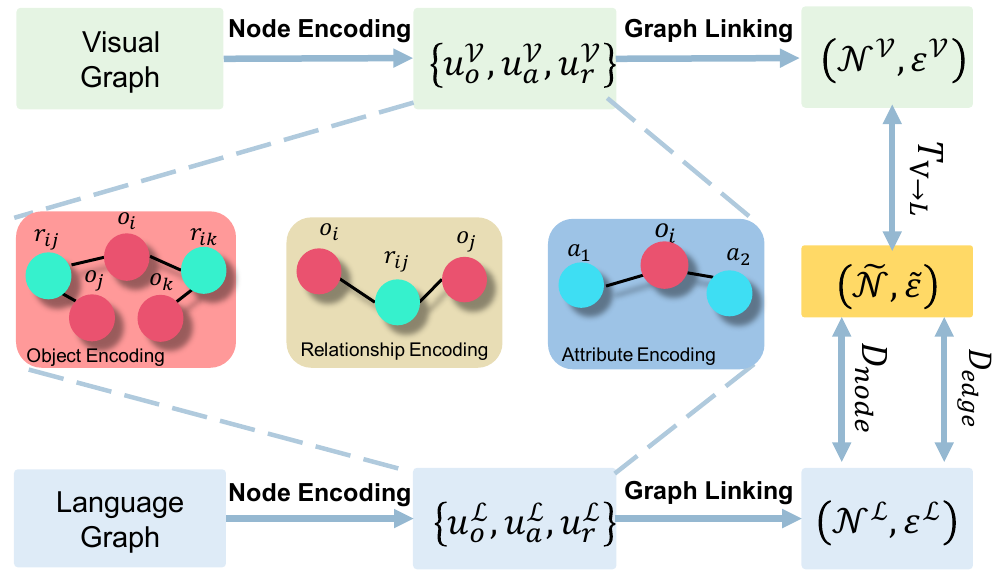}
\centering\caption{An overview of the consensus representation generation. The translator $T_{P\rightarrow K}(\cdot)$ aims to fool the node discriminator $D_{node}$ and edge discriminator $D_{edge}$ through adversarial training.  }
\vspace{-0.3cm}
\end{figure}

\noindent$\textbf{Node Alignment.}$
For node representation alignment, we utilize a node discriminator $D_N$ to discriminate which source the latent vector comes from, and force the representation of $\mathcal{SG^V}$ obey the $\mathcal{SG^L}$ distribution. Given a pair  ($\mathcal{N}_c^{\mathcal{V}},\mathcal{N}_c^{\mathcal{L}}$), the discriminator $D_N$ is trained with the following objective:
\begin{equation}\label{6}
\begin{aligned}
\mathcal{L}(N)&= \mathbb{E}_L[{\rm log} D_N( \mathcal{N}_c^{\mathcal{L}} )] + \phi_{N} \cdot \mathbb{E}_L[{\rm log} (1-D_N( \mathcal{N}_c^{\mathcal{\hat{L}}} )])
 \\ & +  \psi_{N} \cdot \mathbb{E}_V[{\rm log} (1- D_N(T_{V\rightarrow L}( \mathcal{N}_c^{\mathcal{V}}))]
\end{aligned}
\end{equation}
where $\mathcal{U}^{\mathcal{\hat{L}}}$ is the other $\mathcal{SG^L}$ representation from the textual corpus,   $\phi_{N}$ and $\psi_{N}$ are the hyper-parameters.

\noindent$\textbf{Edge Alignment.}$
Similarly, the edge alignment is based on an edge discriminator $D_E$.  As the structural information is
recorded in the edges of structural graph $\mathcal{G}_s$, the alignment of structures can be modeled as the alignment between edges of $\mathcal{E}_{c}^{\mathcal{V}}$ and $\mathcal{E}_{c}^{\mathcal{L}}$. Thus, a consistency loss is designed to align the egdes.
\begin{equation}
\begin{aligned}
\mathcal{L}(E)&= \mathbb{E}_L[{\rm log} D_E( \mathcal{H}(\mathcal{E}_{s}^{\mathcal{L}},\! \mu^{\mathcal{L}}) )] + \phi_{E} \cdot \mathbb{E}_L[{\rm log} (1-D_E \\& (  \mathcal{H}(\mathcal{E}_{s}^{\mathcal{\hat{L}}},\! \mu^{\mathcal{\hat{L}}}  )])
+  \psi_{E} \cdot \mathbb{E}_V[{\rm log} (1- D_E(T_{V\rightarrow L}(   \mathcal{H}(\mathcal{E}_{s}^{\mathcal{V}},\! \mu^{\mathcal{V}}))]
\end{aligned}
\end{equation}
where $\mathcal{H}(\mathcal{E},\mu)=1 /(1+e^{-(\mathcal{E}-\mu)})$ is a logistic function with sigmoid's midpoint $\mu$. $\mu$ is the mean of $\mathcal{E}_{s}$. Constrained by the above Equation, similar edges in $\mathcal{E}_{s}^{\mathcal{V}}$ remain closely  $\mathcal{E}_{s}^{\mathcal{L}}$ located in the feature space.  $\phi_{E}$ and $\psi_{E}$ are the hyper-parameters.

On the basis of the aforementioned graph alignment, each node in the aligned graph ($\tilde{\mathcal{N}}_c,\tilde{\mathcal{E}}_c$) is concatenated by itself and the product of edge and its adjacent nodes. Thus, the consensus representation $\tilde{\mathcal{U}}$=\{$\tilde{\textbf{u}}_o,  \tilde{\textbf{u}}_a, \tilde{\textbf{u}}_r$\} is obtained, and then fed into the captioning module, providing a consensus for image captioning and object grounding.

\subsection{Captioning  Module}
In this module, we exploit the latent information of spatial region proposals and aforementioned consensus representation for caption generation. a

\noindent$\textbf{Augmented Region Proposals.}$
The region proposals $\mathcal{R}= \{\textbf{r}_1, \textbf{r}_2, \cdots, \textbf{r}_N\}$ may contains multiple visual concepts, each $\textbf{r}_i = [x, y, w, h]$ corresponds to a 4-dimensional spatial coordinate, where (x, y) denotes the coordinate of the top-left point of the bounding box, and (h,w) corresponds to the height, width of the box. To exploit their latent spatial relations, following~\cite{yao2018exploring}, we link the proposals by their relative distance to a visual graph $\mathcal{G}_s = (\mathcal{N}_s, \mathcal{E}_s)$.   Except for the two regions are too far away from each other, i.e., their Intersection over Union \textless 0.5, the spatial relation between them is tend to be weak and no edge is established in this case. Finally, the proposals are re-encoded by a GCN, which allows information passing across all proposals.
\begin{equation}
\begin{aligned}
\tilde{\textbf{r}}_i=& \rm \sum_{j =1}^{N_{r_i}} E_{region}(\textbf{r}_i, \textbf{s}_{ij},  \textbf{r}_j )
\end{aligned}
\end{equation}
where $N_{r_j}$ is the total number of the region propsals that associated with $\textbf{r}_i$. $E_{region}(\cdot)$ is the graph convolutional operation for region proposals. Thus, the augmented region proposals  are given by $\tilde{\mathcal{R}}=\{\tilde{\textbf{r}}_1, \tilde{\textbf{r}}_2, \cdots, \tilde{\textbf{r}}_N\}$.

\noindent$\textbf{Sentence Generator.}$
We extend the state-of-the-art of Top-Down visual attention language model~\cite{anderson2018bottom} as our captioning model. This model consists of two-layer LSTMs :  Attention LSTM and Language LSTM (see Fig. 2b). The first one for for encoding the image features $\textbf{v}^{g}$ and embedding of the previously generated word $\textbf{y}_{t-1}$ into the hidden state $\textbf{h}_{t}^{A}$. The second one for caption generation.  We allow the Language LSTM to select dynamically how much from the augmented regions and consensus  to generate words. Specifically, a soft attention is developed on $\tilde{\mathcal{U}}$ and $\tilde{\mathcal{{R}}}$ by $\textbf{h}_{t}^{A}$ from Attention LSTM, and then encoding the attended augmented regions $\hat{\mathcal{{R}}}$ and consensus $\hat{\mathcal{{U}}}$ into into the hidden state $\textbf{h}_{t}^{L}$. In this setting, the captioning model can select the relevant information it needs on the basis of the current semantic context to describe an image reasonably.

Using the notation $\textbf{y}_{1:T}$ is refer to a sequence of words $\{\textbf{y}_1, \cdots, \textbf{y}_T \}$. Each step $t$, the conditional distribution over possible words is generated by the Softmax, the cross-entropy loss  $\mathcal{L}(S)$ for captioning can be computed by:
\begin{equation}\label{10}
\mathcal{L}(S) =  -\lambda_L \cdot \sum^T_{i=0} log(p(\textbf{y}_t| \textbf{y}_{0:T-1}))
\end{equation}
%

\subsection{Grounding Module}
In this section, we aim to evaluate how well the captioning model grounds the visual objects.

\noindent$\textbf{Region Grounding.}$  To further assist the language model to attend to the correct regions, following ~\cite{zhou2019grounded}, we develop the region attention loss $ \mathcal{L}(R)$: we denote the indicators of positive/negative regions as $\rho = \{\rho_1, \cdots, \rho_N\}$ in each time step, where $\rho_i =1$ if the region $\rho_i$ has over 0.5 IoU with GT box and otherwise 0. Combining the treating attention $a^r$ (in section of Captioning Module), the proposal attention loss function is defined as:
\begin{equation}\label{11}
\mathcal{L}(R)= -\lambda_R \cdot \sum _1^N \gamma_i log a_i^r
\end{equation}

\noindent$\textbf{Object Localizaiton.}$ Given a object word $\textbf{w}$ with a specific class label, we aim to localize the related region proposals. We first define the region-class similarity function with the treating attention weights $a^p$ as below:
 \begin{equation}
\begin{aligned}
p^s(\textbf{r}, a^r) &=  {\rm Softmax} ({W}_s ^ \top \textbf{r}  + a^r )
\end{aligned}
\end{equation}
where $W_s$ $\in \mathbb{R}^{ d \times N}$ is a simple object classifier to estimate the class probability distribution.

Thus we use the $p^s(\textbf{r}, a^r)$ to calculate the confidence score for $\textbf{w}$, the grounding loss function $\mathcal{L}(G)$ for word $\textbf{w}$ is defined as follows:
\begin{equation}\label{13}
\mathcal{L}(L)= -\lambda_L \cdot \sum ^N_{i=1}\gamma_i log  p^s (\textbf{r}_i)
\end{equation}

\subsection{Training Algorithm}
Algorithm 1 (see in appendix) details the pseudocode of our CGRL algorithm for GIC. First, we pretrain the language GCNs by reconstructing the sentences from the latent vector that is encoded from the language graph $\mathcal{SG}^{\mathcal{L}}$ with the GCN (sentence $\rightarrow$ $\mathcal{SG}^{\mathcal{L}}$ $\rightarrow$ latent vector $\rightarrow$ sentence $\bar{\mathcal{S}}$  $\rightarrow$ object grounding $\bar{\mathcal{R}}^o$). Then, we keep the language GCN fixed, and learn the visual GCN by aligning the visual graph $\mathcal{SG}^{\mathcal{V}}$   to the language graph $\mathcal{SG}^{\mathcal{L}}$   in the latent space. Basically, the encoded latent vector from the knowledge $\mathcal{SG}^{\mathcal{L}}$  is used as supervised signals to learn the visual GCN.

\begin{table*}[t]
  \centering
    \begin{tabular}{l |c| c c c c c| c c c c}
    \hline
     &\multicolumn{1}{|c|}{} &\multicolumn{5}{c|}{ Captioning Evaluation } &\multicolumn{4}{c}{ Grounding Evaluation }\\
    \hline
    Method & CR. & BLEU@1  & BLEU@4   & METEOR   & CIDEr  & SPICE  & GRD. & ATT.  & F1$_{ALL}$ & F1$_{LOC}$\\
    \hline
    \hline
    NBK~$^{\ast}$                             && 69.0 &27.1 &21.7& 57.5 &15.6 & - & - & - &-\\
    Cyclical~$^{\ast}$                  & & 68.9& 26.6 &22.3 &60.9 &16.3&-&-&4.85& 13.4\\
    \hline
    GVD$^{\dagger}$       &&69.9& 27.3& \textbf{22.5} &62.3 &16.5& 41.4 &50.9 &7.55 &22.2 \\
    \hline
    CGRL (w/o CR)$^{\ddagger}$       &  \checkmark & 70.0 &27.4& 22.4& 62.9 &16.5 &41.6 &51.0& 7.48 &22.1\\
     CGRL (w/o ARP)$^{\ddagger}$     &\checkmark  & 69.9& 27.0& 22.2& 61.2& 16.3& 27.4& 29.5 &4.49& 12.8\\
    CGRL (w/o OG)$^{\ddagger}$       &\checkmark  & \textbf{72.9} &\textbf{28.3}& 22.4& \textbf{65.4} &\textbf{16.8} &\textbf{45.5} &\textbf{55.9}& - &-\\
    CGRL (w/o NA)$^{\ddagger}$                               &\checkmark & 70.9 &26.8& 21.3& 62.1 &16.3 &41.9 &50.7& 7.59 &22.6\\
    CGRL (w/o EA)$^{\ddagger}$ &\checkmark  & 72.7 &27.4& 22.1& 64.1 &16.8 &43.9 &54.1& 8.00 &23.4\\
    \hline
     CGRL$^{\ddagger}$                                          &\checkmark & 72.5 &27.8& 22.4& 65.2 &\textbf{16.8} &44.3 &54.2&\textbf{8.01} &\textbf{23.7}\\
    \hline
    \end{tabular}
    \caption{Captioning results and grounding results on Flickr30k Entities test set.  CR. indicates the consensus  representation  knowledge is used or not for GVD generation.  $\ast$ indicates the results are obtained from the original papers. $^{\ddagger}$ captions are  implemented by author form the code of original paper. $^{\ddagger}$ sentences obtained directly from the author. Larger value indicates better performance. Top one score on each metric are in bold. Acronym notations of each method see in Sec 4.3.}
    \vspace{-0.3cm}
\end{table*}

\begin{table}[t]
  \centering
    \begin{tabular}{l |c|c c c  }
    \hline
     & \multicolumn{1}{|c|}{CR.} & \multicolumn{3}{c}{ Hallucination Evaluation}\\
    \hline
    Method  && CHAIR$_{i}$ & CHAIR$_{s}$ & RECALL$_o$  \\
    \hline
    \hline
   CGRL (w/o CR)$^{\ddagger}$      && 0.437 & 0.732& 0.541 \\
     CGRL (w/o ARP)$^{\ddagger}$    & \checkmark& 0.385 & 0.718& 0.573 \\
    CGRL (w/o NA)$^{\ddagger}$     & \checkmark& 0.392 & 0.718& 0.567\\
     CGRL (w/o EA)$^{\ddagger}$    &\checkmark & 0.361& 0.712& 0.598 \\
     CGRL (w/o OG)$^{\ddagger}$     & \checkmark& 0.379 & 0.712& 0.629\\
     \hline
     CGRL$^{\ddagger}$              &\checkmark& \textbf{0.347} & \textbf{0.707}& \textbf{0.644} \\
    \hline
    \end{tabular}
    \caption{ Hallucination results on Flickr30k Entities test set. Lower value of CHAIR$_{i}$, CHAIR$_{s}$ indicates better performance, while the large value of RECALL$_o$ is better. }
\vspace{-0.3cm}
\end{table}

\section{Experiments}
\subsection{Dataset and Setting}
\noindent$\textbf{Dataset.}$ We benchmark our approach for GIC on the Flickr30k Entities dataset and compare our CRGL method to the state-of-the-art models. Moreover, the Flickr30k Entities collected 31k images with 275k bounding boxes with associated with natural language phrases. Each image is annotated with 5 crowdsourced captions.
There are 290k images for training, 1k images for validation, and another 1k images for testing.

\noindent$\textbf{Settings.}$  Given an image  as the input, the region encoding $r$ is extracted from  pre-trained Faster R-CNN~\cite{ren2015faster}.  The  global visual feature $\bar{v}^{g}$  contains its all-region features which are given by ResNeXt-101~\cite{xie2017aggregated}. For image captioning, we tokenized the texts on white space, and the sentences are ``cut'' at a maximum length of 20 words. All the Arabic numerals are converted to the English word. We add a \emph{Unknown token} to replace the words out of the vocabulary list. The vocabulary has  7,000 words, and each word is represented by a 512-dimensional vector, the RNN encoding size m = 1024.
\subsection{Comparing Methods} We compare the proposed Consensus Graph Representation Learning (CGRL) algorithm with existing SoTA method NBK~\cite{lu2018neural}, Cyclical~\cite{ma2019learning} and GVD~\cite{zhou2019grounded}  on Flickr30k Entities dataset.

 We also conduct an ablation study to investigate the contributions of individual components in CGRL. In our experiment, we train the following variants of CGRL:
\textbf{CGRL (w/o CR)}, which is a general GIC model that only use image features to generate grounded image caption. \textbf{CGRL (w/o ARP)}, generating captions without augmented region proposals to study the importance of region supervision. \textbf{CGRL (w/o OG)}, generating captions sequentially without object word grounding, which is similar to standard image captioning algorithm. \textbf{CGRL (w/o NA)}, generating captions without node alignment in Generative Adversarial Structure Network. \textbf{CGRL (w/o EA)}, generating captions without edge alignment in Generative Adversarial Structure Network.
\subsection{Metrics}
\noindent \textbf{Automatic Metrics.} We use the captioning evaluation tool  provided by the 2018 ActivityNet Captions Challenge, which includes \emph{BLEU@1}, \emph{BLEU@4}, \emph{METEOR}, \emph{CIDEr}, and \emph{SPICE} to evaluate the captioning results. Following \footnote{https://github.com/facebookresearch/ActivityNet-Entities}, we compute Fl$_{ALL}$, Fl$_{LOC}$, \emph{GRD.} and \emph{ATT.} to  evaluate the grounding results. For hallucination evaluation, we compute the   CHAIR$_i$ and CHAIR$_s$  provide by ~\cite{rohrbach2018object}. Besides, we also record the recall of object words  $RECALL_o$ that correctly predicted in each sentence. See the appendix for more details.

\begin{figure*}[t]
\includegraphics[width=0.95\textwidth]{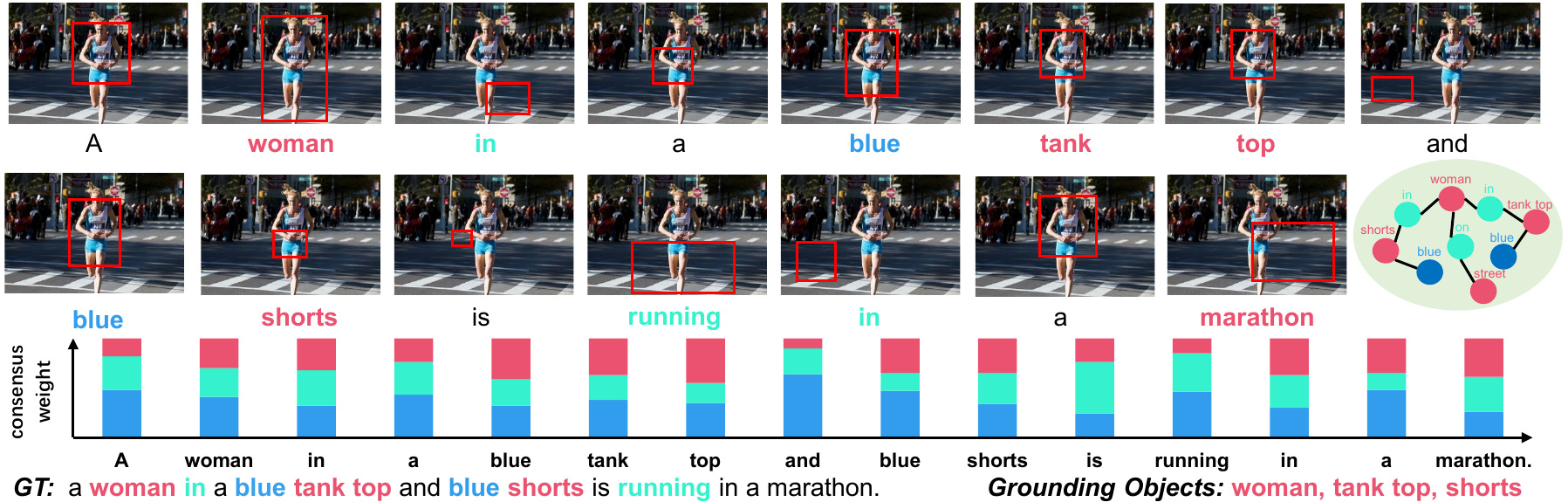}
\centering\caption{Visualization of image caption generation. Red box corresponds to the region with the highest attention $a^r$. Histogram presents the consensus weight for $\tilde{\textbf{u}}_o, \tilde{\textbf{u}}_a, \tilde{\textbf{u}}_r$, respectively.}
\vspace{-0.3cm}
\end{figure*}

\noindent \textbf{Human Evaluation.} To better understand how satisfactory are the sentences generated from our CGRL, we also conducted a human study with 5 experienced workers to compare the descriptions generated by CGRL and CGRL (w/o CR), and asked them which one is more descriptive and captures more key visual concepts. Besides, we validate relevance of object/relationship/attribute words in generated captions by R$_{object}$, R$_{relationship}$  and R$_{attribute}$. On the other hand, we also allow humans to evaluate which caption is more descriptive subjectively by $DES$. For each pairwise comparison, 100 images are randomly sampled from the Karpathy split for the workers to evaluate.

\subsection{Experimental Results}

\noindent$\textbf{Captioning Results.}$ Table~1 shows the overall qualitative results of our model and SoTA on the test set of Flickr30k Entities dataset. In general, CGRL achieves the best performance on almost all the metrics to SoTA, METEOR  also gets a comparable result with a small gap (within 0.1). Notice that we obtain 0.6 improvements on SPICE, since our method learned the consensus representation of $\mathcal{SG}$, which gives a positional semantic prior to improve this score.

\noindent$\textbf{Grounding Results.}$ Table~1 also presents our method effectively improves the accuracy of GRD., ATT., F1$_{ALL}$ and F1$_{LOC}$. According to our observation, CGRL almost obtains the best performance for the majority of grounding metrics (except for ATT.). Since the captioning benefit from the consensus representation and grounding regions to generate correct words.  Notice that CGRL (w/o OG) outperform other variants on attention correctness ATT, our intuition is that the CGRL (w/o OG) without object grounding may pay more attention to the word localization on GT sentence.

 \newcommand{\tabincell}[2]{\begin{tabular}{@{}#1@{}}#2\end{tabular}}
\begin{table}[t]
  \centering
    \begin{tabular}{ l| c c c c c}
    \hline
    & \multicolumn{3}{c}{ Human Evaluation}\\
     \hline
    Metric &      \tabincell{c}{CGRL is Better  }   & \tabincell{c}{ CGRL is  Worse }  & Equal \\
    \hline
    \hline
    R$_{object}$ \rule{0pt}{10pt}&\textbf{0.26 (+5\%)} &0.21&0.53  \\
    R$_{relationship}$ \rule{0pt}{10pt}&\textbf{0.35 (+12\%)}& 0.23& 0.42 \\
    R$_{attribute}$ \rule{0pt}{10pt}&\textbf{0.17 (+3\%)} &0.14&0.69  \\
    \hline
    DES \rule{0pt}{10pt}&\textbf{0.22 (+6\%)}&0.16&0.62  \\
    \hline
    \end{tabular}
    \caption{ Human evaluation of our model and its variant CGRL (w/o CR) on Flickr30k Entities test set. For each pairwise comparison are compared by 100 random images. Larger value indicates better performance. }
\vspace{-0.3cm}
\end{table}
\begin{figure}[t]
\includegraphics[width=0.475\textwidth]{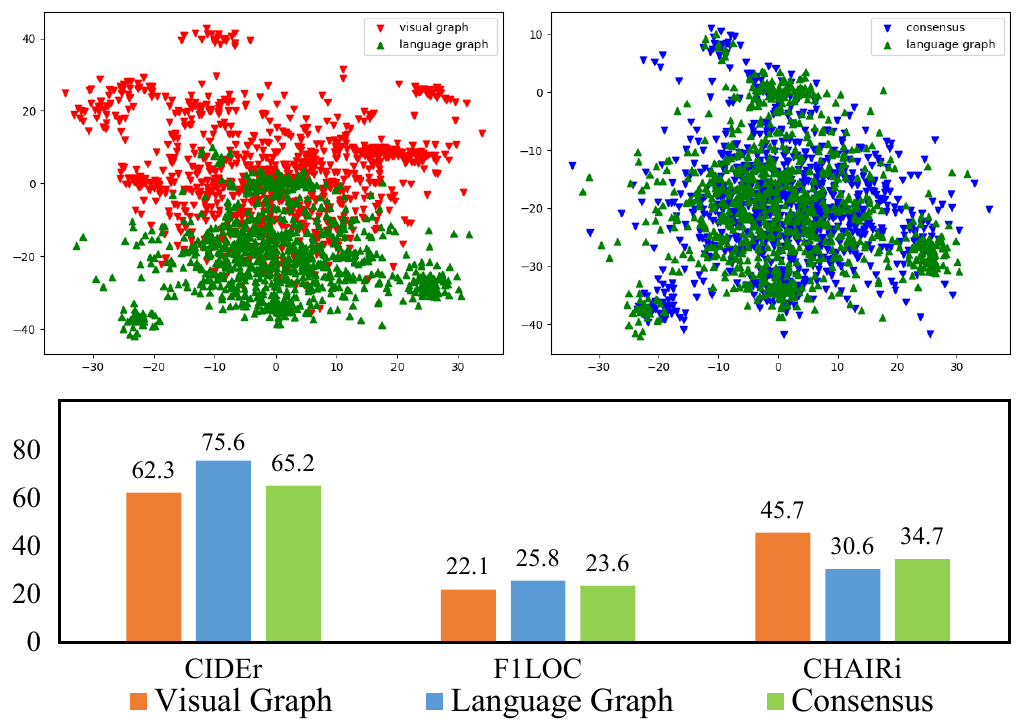}
\centering\caption{Visualization of features (1000 cases) in 2D space by t-SNE and the GIC performance of different knowledge. To facilitate the comparison, the value of CHAIR$_i$ is multiplied by 100.}
\vspace{-0.5cm}
\end{figure}

\noindent$\textbf{Hallucination Results.}$  Table~2 presents object hallucination on the test set. We note an interesting phenomenon that the consensus based methods tend to perform better on the CHAIR$_{i}$, CHAIR$_{s}$ and RECALL$_o$ metrics than methods without consensus  by a large margin. This proves the consensus contains key knowledge that can help the captioning model capture more correct objects in image. What's more, based on this basis, the hallucination of objects is decreased, we believe that the consensus  can assist the region grounding operation in generating correct object words.

\noindent$\textbf{Human Evaluation.}$  As commonly known, the text-matching-based metrics are not perfect, and some descriptions with lower scores actually depict the images more accurately, i.e., some captioning models can describe an image in more details but with the lower captioning scores. Thus, we conducted a human evaluation.   The results of the comparisons are shown in Table~3. It is seen from the table that CGRL outperforms CGRL (w/o CR) in terms of R$_{object}$, R$_{relationship}$  and Rt$_{attribute}$ by a large margin. The results indicate that our model with consensus can capture more key visual concepts in generated captions.  This proves the effectiveness of CGRL for grounded image captioning. On another side, the sentence produced by CGRL achieves a higher score of the $DES$ evaluation. We believe this result benefits from the consensus in CGRL.

\subsection{Quantitative Analysis}
We further validate the several key issues of the proposed method by answering three questions as follows.

$\textbf{Q1: How much improvement of semantic inconsistency}$\\
$\textbf{ has the consensus brought?}$
First, the visualization of features alignment indicates that our GASN is able to capture the semantic consistency across vision-language domain in Figure~4. To justify the contribution of consensus, we investigated the importance of different knowledge by three variants of CGRL. In Figure~4, the best performance is to use the $\mathcal{SG}^{\mathcal{L}}$  directly. This is reasonable since the $\mathcal{SG}^{\mathcal{L}}$  contains the key concepts to reconstruct its description naturally.  In addition,  consensus-based outperforms $\mathcal{SG}^{\mathcal{V}}$  by a large margin on all the important metrics, especially the CHAIRi (-11.0).  The experimental results indicate that the consensus can improve the quality of GIC greatly.


$\textbf{Q2: How does the CGRL generate the captions based}$\\
$\textbf{on consensus and region attention?}$\\We visualize the process of caption generation in Figure~8(a)(In appendix). On the one hand, from the the histogram, it clearly demonstrates the CGRL select \emph{how much} from the consensus $\tilde{\textbf{u}}_o, \tilde{\textbf{u}}_a, \tilde{\textbf{u}}_r$, respectively. For example, consensus knowledge is more important while generating the relational words ``\emph{run}''. On the other hand, CGRL attends to the appropriate proposal to predict a word, such the word ``\emph{woman}''. In summary,  on the basis of the consensus attention and region attention, our model can select the concerned information to generate the correct word.


\section{Conclusion}
In this paper, we propose a  novel Consensus Graph Representation Learning (CGRL) framework to train a GIC model. We design a consensus based method that aligns the visual graph to language graph through a novel generative adversarial structure network, which aims to maintain the semantic consistency between multi-modals.  Based on the consensus, our model not only can generate more accurate caption, but also ground appropriate regions and greatly alleviate the hallucination problem.  We hope our CGRL can provide a complement for existing literature of visual captioning and benefit further study of vision and language.

 \section{ Acknowledgments}This work has been supported in Apart by National Key Research and Development Program of China (2018AAA010010), NSFC (U19B2043, 61976185), Zhejiang NSF (LR21F020004, LR19F020002), Key Research \& Development Project of Zhejiang Province (2018C03055), Funds from City Cloud Technology (China) Co. Ltd., Zhejiang University iFLYTEK Joint Research Center, Zhejiang University-Tongdun Technology Joint Laboratory of Artificial Intelligence, Chinese Knowledge Center of Engineering Science and Technology (CKCEST), Hikvision-Zhejiang University Joint Research Center, the Fundamental Research Funds for the Central Universities, and Engineering Research Center of Digital Library, Ministry of Education.

\bibliographystyle{aaai21}
\bibliography{ref}
\end{document}